\documentclass[final]{cvpr}

\usepackage{times}
\usepackage{epsfig}
\usepackage{graphicx}
\usepackage{amsmath}
\usepackage{amssymb}
\usepackage{algorithm}
\usepackage{algorithmic}
\usepackage{stfloats}

\usepackage[pagebackref=true,breaklinks=true,colorlinks,bookmarks=false]{hyperref}

\def\cvprPaperID{****} 
\def\confYear{CVPR 2021}

\begin{document}

\title{2nd Place Solution for 

Waymo Open Dataset Challenge --- Real-time 2D Object Detection}

\author{Yueming Zhang\textsuperscript{1,2} ~~~~ Xiaolin Song\textsuperscript{1,2} ~~~~ Bing Bai\textsuperscript{2} ~~~~ Tengfei Xing\textsuperscript{2}  ~~~~ Chao Liu\textsuperscript{2} ~~~~ Xin Gao\textsuperscript{2}  \\
Zhihui Wang\textsuperscript{2} ~~~~ Yawei Wen\textsuperscript{2} ~~~~ Haojin Liao\textsuperscript{2} ~~~~ Guoshan Zhang\textsuperscript{1} ~~~~ Pengfei Xu\textsuperscript{2}  \\
1.Tianjin University ~~~~~~~~  2.Didi Chuxing\\
{\tt\small seife@tju.edu.cn} \    {\tt\small xupengfeipf@didiglobal.com}
}

\maketitle

\begin{abstract}
In an autonomous driving system, it is essential to recognize vehicles, pedestrians and cyclists from images. Besides the high accuracy of the prediction, the requirement of real-time running brings new challenges for convolutional network models. In this report, we introduce a real-time method to detect the 2D objects from images. We aggregate several popular one-stage object detectors and train the models of variety input strategies independently, to yield better performance for accurate multi-scale detection of each category, especially for small objects. For model acceleration, we leverage TensorRT to optimize the inference time of our detection pipeline. As shown in the leaderboard, our proposed detection framework ranks the 2nd place with 75.00\% L1 mAP and 69.72\% L2 mAP in the real-time 2D detection track of the Waymo Open Dataset Challenges, while our framework achieves the latency of 45.8ms/frame on an Nvidia Tesla V100 GPU.
\end{abstract}

\begin{figure*}[!ht]
    \centering
    \includegraphics[width=0.9\textwidth]{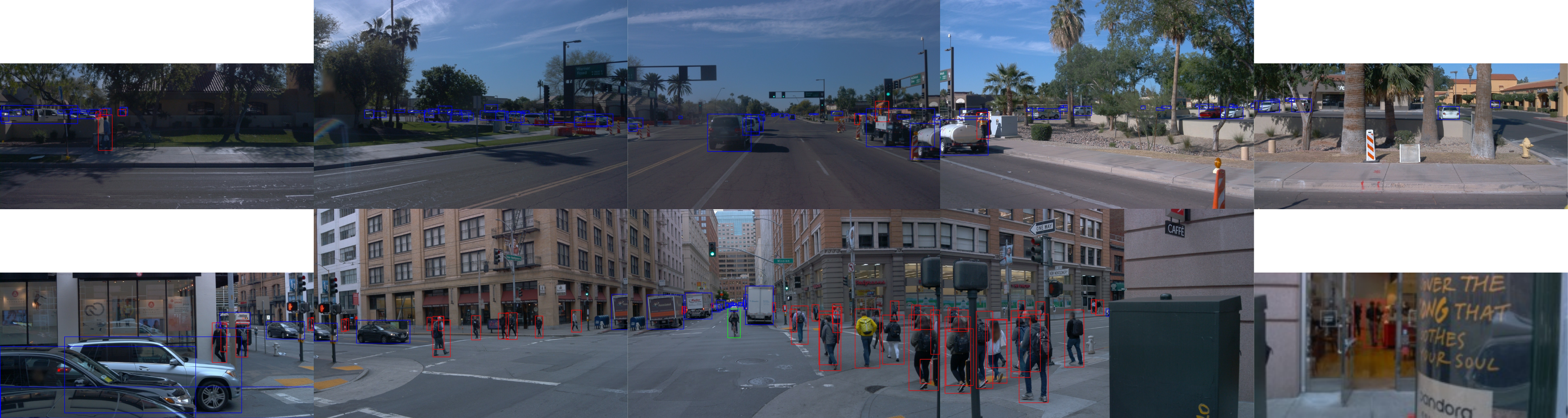}
    \caption{Examples of images in Waymo Open Dataset, there are totally 5 cameras with different resolution.}
    \label{waymo}
\end{figure*}

\section{Introduction}

It is practical to recognize different objects on the road for an autonomous driving system, especially object detection tasks are the basic solutions for images captured by cameras on the public roads. The Waymo Open Dataset challenges provide a platform for different scenes. The Waymo Open Dataset \cite{sun2020scalability} collects the labeled images captured from high-quality cameras and LiDAR sensors which are utilized on the vehicle for multi-view modeling in real autonomous driving. In real-time 2D detection task, there are three categories (vehicle, pedestrian and cyclist) annotated with 2D bounding boxes. Compared with the 2D detection challenge in 2020, the real-time 2D detection challenge requests that the submitted model must run faster than 70ms/frame on a Nvidia Tesla V100 GPU in this year.
We conduct a state-of-the-art real-time 2D object detection framework in this challenge.

\section{Our Solution}

\subsection{Base Detectors}

Deep learning based object detectors are roughly divided into two mainstreams, one-stage detector and two-stage detector. In order to achieve robust detection results, we aggregate the popular one-stage object detectors with different settings. Considering of the runtime, we choose the popular one-stage detector YOLOR \cite{wang2021you}, which is the one of the upgraded version from YOLOv1 \cite{redmon2016you}, and achieves the state-of-the-art of real-time object detection on COCO.

\subsubsection{YOLO}

The core idea of YOLOv1 is to use the whole graph as the input of the network and directly return to the location and category of the bounding box in the output layer. Compared with other object detection algorithms, the detection speed of YOLO is very fast. In YOLOv2\cite{redmon2017yolo9000}, Some new features have been added such as batch normalization, high resolution classifier, convolutional with anchor boxes, dimension clusters and Darknet-19 network. YOLOv3\cite{redmon2018yolov3} uses Darknet-53 as the backbone and uses the idea of pyramid feature map for reference, small feature map is used to detect large objects, while large feature map is used to detect small objects. The main innovations of YOLOv4\cite{sun2020scalability} are mosaic data enhancement, self-adversarial training and cross mini-batch normalization.

\subsubsection{YOLOR}
YOLOR applies the implicit and explicit knowledge to the model training at the same time so that it can learn a general representation and complete various tasks through this general representation. The implicit knowledge can be modeled by vector $z$, neural network $Wz$, or matrix factorization $Z^Tc$. Through kernel space alignment, prediction refinement, and multi-task learning, implicit knowledge is integrated into the explicit knowledge which can greatly improve the performance and generalization ability of the model. Compared with other detection methods on COCO dataset, the mAP of YOLOR is 3.8$\%$ higher than PP-YOLOv2\cite{huang2021pp} at the same inference speed and the inference speed has been increased 88$\%$ at the same accuracy compared with Scaled-YOLOv4-P7\cite{wang2020scaled}. We choose YOLOR as our detector.


\begin{figure*}[t]
    \centering
    \includegraphics[width=0.9\textwidth]{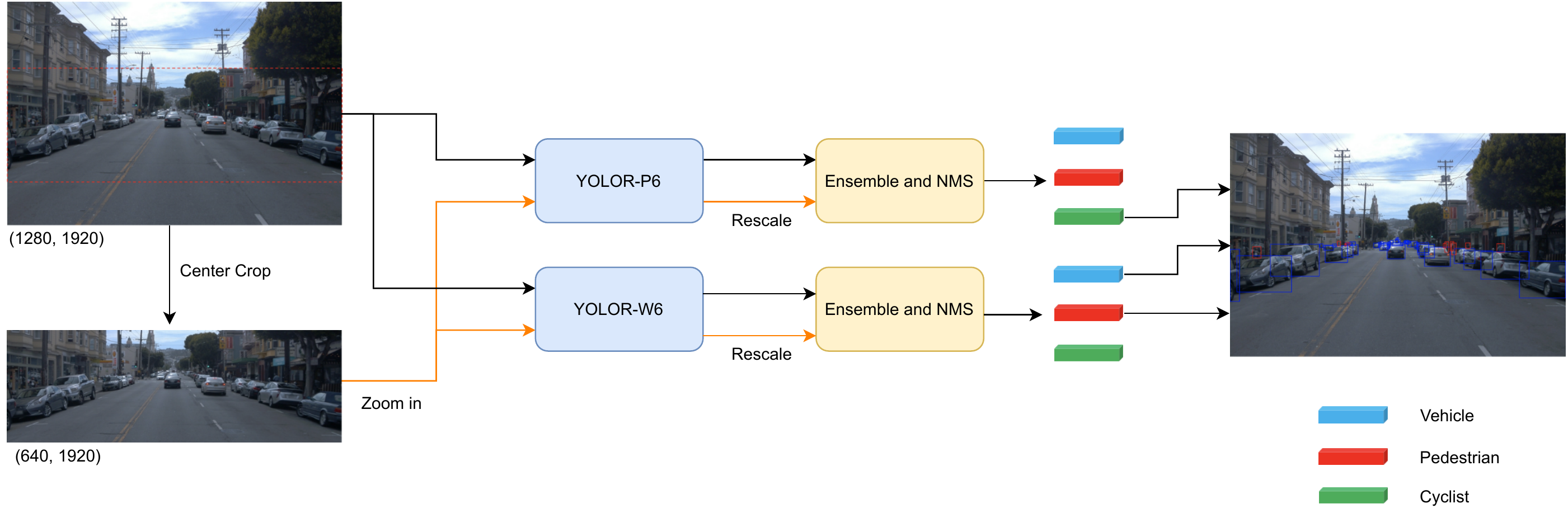}
    \caption{The pipeline of our solution.}
    \label{small object}
\end{figure*}
\subsection{Scale Enhancement for Small Objects}
\label{CRE}
\begin{figure}
    \centering
    \includegraphics[width=0.7\columnwidth]{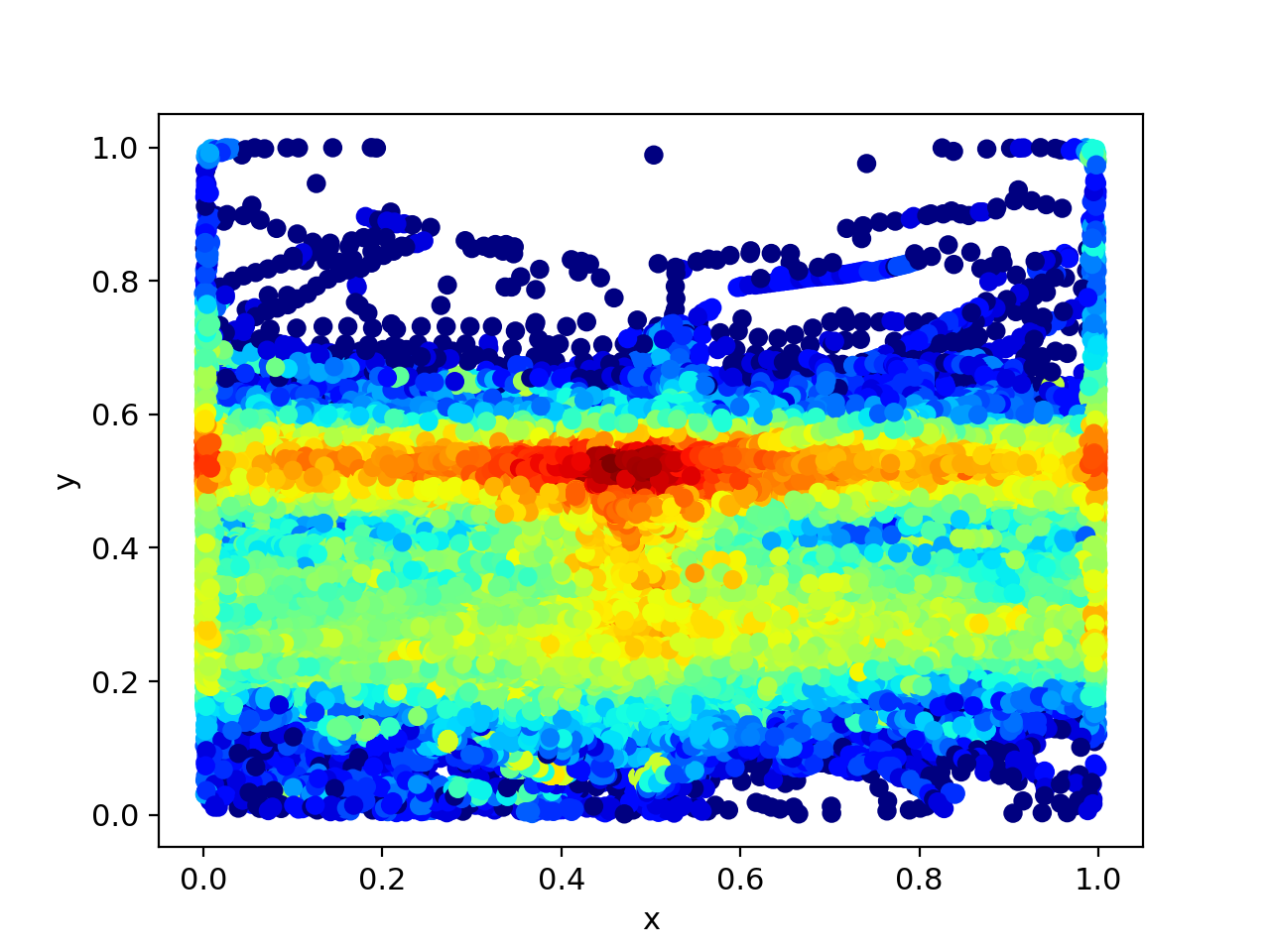}
    \caption{Statistics heatmap of small objects' center in the training dataset.}
    \label{heatmap}
\end{figure}

According to the statistical distribution of the waymo-open-dataset. Nearly one-third of objects belong to small objects, whose areas are less than 32$\times$32 pixels. Due to the fixed camera position in waymo data-collection car, the scene of the target is relatively fixed. The top-left image in Figure \ref{small object} is a typical example. The small vehicle objects always locate at the mid part of the image. And we make a statistic on this attribute. As shown in Figure \ref{heatmap}, most of the small box centers locate at the center area of the y-axis. So, we crop the center area and resize it into 1.5$\times$ larger. This strategy can help with detecting these small objects. As shown in Figure \ref{small object}, given an image, we first use the trained model to predict the bounding-box in original shape (1920$\times$1280 or 1920$\times$886), which we called Result A. Next, we crop the target area ($0\times W$, $0.3\times H$, $1.0\times W$, $0.8\times H$) and resize the cropped image into 1.5$\times$scale (2880$\times$960 or 2880$\times$704). Then we input the enlarged image to the same model to get an enlarged result and rescale it into the original size for the results ensemble, which is called Result B. Because we use the rescaled image to make the finest predictions on small objects. So, we remove all targets whose areas are larger than 96$\times$96 in Result B. Then the Result A and the bounding-boxes whose areas are less than 96$\times$96 in Result B are collected together for non-maximum-suppression. 

\subsection{Anchor Selection}
We use K-Means clustering to generate the hyperparameters for Real-time 2D Detection on the Waymo Open Dataset. However, the model fails to outperform baseline using our hyperparameter setting, as shown in Table \ref{anchorselection}. It may possibly result from the fact that the width and height of small anchor set changes a lot after clustering with medium and large objects, e.g. the aspect ratio of the smallest anchor is 1.07, but it is 0.703 in original set. To validate our conjecture, we only do clustering for small objects ($W\times H<64\times 64$) and achieve better results for vehicle and pedestrian but worse results for cyclist, where vehicle and pedestrian involve more small objects than cyclist. Since the AP of vehicle and pedestrian is much lower than YOLO-W6 in Table \ref{P6vsW6}, and only the cyclist result of YOLO-P6 is adopted in final model ensembling, we adopt baseline result of YOLO-P6 to get better results.

\begin{table}[h]
\begin{center}
\caption{Anchor Selection Experiment. Veh., Ped. and Cyc. refer to Vehicle, Pedestrian and Cyclist, correspondingly.}
\begin{tabular}{l|c|c|c|c}
\hline
Method  & Veh. & Ped. & Cyc. & Mean \\
\hline
Baseline & 61.5 & 73.5 & 57.3 & 64.1  \\
All Objects Cluster & 61.5 & 74.1 & 56.2 & 63.9 \\
Small Objects Cluster & 61.7 & 74.6 & 55.1 & 63.8\\
\hline
\end{tabular}
\label{anchorselection}
\end{center}
\end{table}

\subsection{Model Acceleration}
Real-time 2D Detection means our model not only needs to achieve accurate detection but also runs at a fast inference time. Although the YOLOR-P6 \cite{wang2021you} can meet this demand, we still need to accelerate the inference time for the multi-model ensemble. To this end, we adopt TensorRT for model inference accelerating. TensorRT is a high-performance deep learning inference optimizer and can greatly reduce the inference time while reserving the accuracy. Table 2 is our experiment result on the mini-val dataset, all results are evaluated at the resolution of (1920, 886). Besides, it supports FP16 precision inference on Nvidia Tesla V100 GPU (16GB), which can further accelerate the inference process and reduce GPU memory cost. The inference time of FP32 precision on Tesla V100 GPU is 37ms while the inference time of FP16 precision achieves 19ms. 

\begin{table}[h]
\begin{center}
\caption{TensorRT Experiments on Nvidia Tesla P40 GPU.}
\begin{tabular}{l|c|c|c|c}
\hline
Method  & Veh. & Ped. & Cyc. & Latency (s) \\
\hline
Baseline & 61.5 & 73.5 & 57.3 &  0.085 \\
With TensorRT & 61.5 & 74.1 & 56.2 & 0.051 \\
\hline
\end{tabular}
\label{tensorrtresult}
\end{center}
\end{table}

\begin{figure}
    \centering
    \includegraphics[width=0.7\columnwidth]{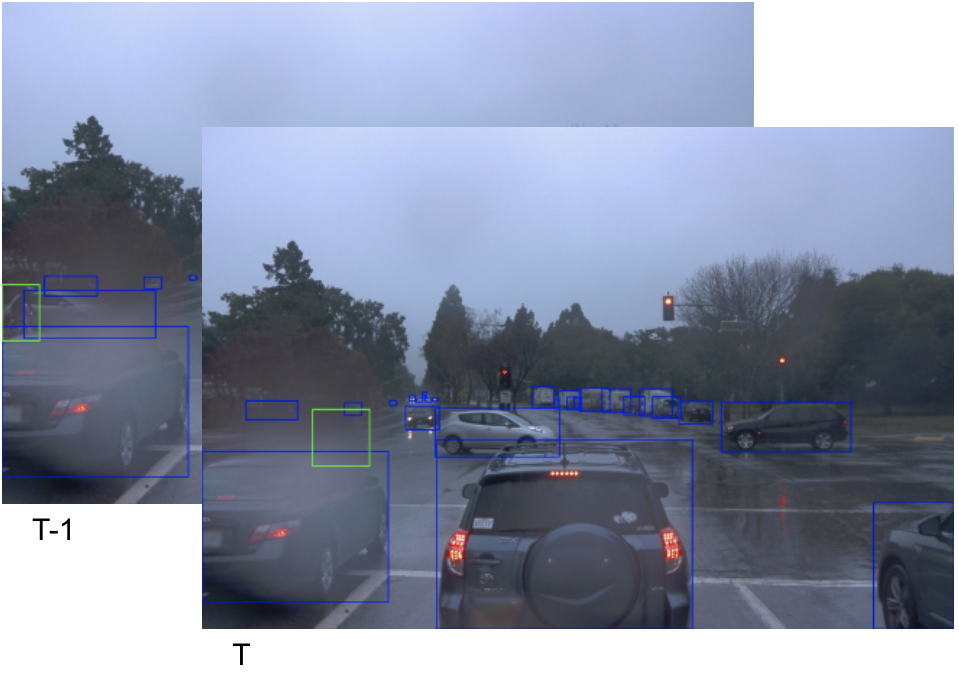}
    \caption{Difficult annotations are colored in green.}
    \label{error}
\end{figure}
\subsection{Training Data Filtering}
When we analyze the current data annotations, we find that there are some difficult annotations. Some objects are occluded, but they are still annotated since the former or the latter frame could indicate its class. As shown in Figure \ref{error}, the green box is occluded in $T$ frame, but it is still labeled as cyclist according to the $T-1$ frame. These annotations are valuable for tracking but bring challenges for the 2D detection method which does not use temporal information. Because these difficult objects may cause problems for the training. Therefore, we learn from the self-learning method which use multi-different models check with each other to automatically clean the dataset during the model training process to solve this problem and improve the model performance.


\section{Experiments}

\subsection{Dataset and Evaluation}

\noindent\textbf{Dataset.}  The Real-time 2D Detection for Waymo Open Dataset \cite{waymowebsite} contains 798, 202 and 150 video sequences in the training, validation, and testing sets, respectively. Each sequence has 5 views, where each camera captures 171-200 frames with the image resolution of 1920 × 1280 pixels or 1920 × 886 pixels. Our models are pre-trained on the COCO dataset then fine-tuned on the Real-time 2D Detection for Waymo Open Dataset. We follow \cite{chen20202nd} and sample the dataset to construct a small dataset that can represent the entire dataset for quick experiment. In the final submission, we use the full dataset to train the model to get better performance.

\noindent\textbf{Evaluation Metrics.}  According to the Real-time 2D Detection for Waymo Open Dataset track, we report detection results on the Level 2 Average Precision (AP) that averages over vehicle, pedestrian, and cyclist classes. The positive IoU thresholds are set to 0.7, 0.5, and 0.5 for evaluating vehicles, cyclists, and pedestrians, respectively. In this track, another important evaluation indicator is speed, which is limited to 70ms/frame on Nvidia Tesla V100 GPU.

\subsection{Implementation Details}
YOLOR-P6 and YOLOR-W6 are two different variants of YOLOR network, both of them use the same down-sampling method, and YOLOR-W6 is a wider version than YOLOR-P6. The base channel number of YOLOR-P6 is \{128, 256, 384, 512, 640\}, while that of YOLOR-W6 is \{128, 256, 512, 768, 1024\}.

\noindent\textbf{YOLOR-P6}.  Cause YOLOR-P6 Network is the fastest network variant in YOLOR series \cite{wang2021you}. We choose YOLOR-P6 as our baseline for meeting the demand of latency and train a 3-class model directly. We fine-tune the pre-trained model in Waymo Open Dataset. To speed up the training process, we use the same data sampling strategy like \cite{chen20202nd}. All models are trained for 50 epochs with a warm-up strategy from 4e-4 to 1e-2. Then the learning rate drops from 1e-2 to 1e-3 in a cosine function. We also use multi-scale training, and the input size multiplies a random coefficient range from $(0.5, 1.2)$. During testing, the model makes predictions on the input at $(1920, 1280)$, which is the original shape of the front camera image. For a side camera image whose resolution is $(1920, 886)$, a constant value is used to pad the image. We use random data augmentation to enhance the data, which includes random flip left-right, random Hue, random Saturation, and image translation. Finally, a class-aware NMS is applied to filter out the overlapped boxes. Different class has different NMS IoU threshold, the setting used in our experiment is $\{'vehicle':0.75, 'pedestrian':0.55, 'cyclist':0.55\}$.

\begin{table*}[!ht]
\begin{center}
\caption{Leaderboard of the Waymo Open Dataset Challenge --- Real-time 2D Detection track \cite{waymowebsite}. The top-5 methods are listed here. The top-2 results are highlighted in red and blue colors, respectively.}
\begin{tabular}{l|c|c|c}
\hline
Method & AP/L1 (\%) & AP/L2 (\%) &Latency (s)\\
\hline
LeapMotor\_Det & \textcolor{red}{0.7571} & \textcolor{red}{0.7041} & \textcolor{red}{0.0616} \\
YOLOR\_TensorRT (Ours) & \textcolor{blue}{0.7500} & \textcolor{blue}{0.6972} & \textcolor{blue}{0.0458} \\
YOLOR\_P6\_TRT & 0.7484 & 0.6956 & 0.0374\\
dereyly\_self\_ensemble & 0.7173 & 0.6565 & 0.0687 \\
YOLO\_v5 & 0.7025 & 0.6414 & 0.0381 \\
\hline
\end{tabular}
\label{tab:leaderboard}
\end{center}
\end{table*}

\noindent\textbf{YOLOR-W6}. Compared with YOLOR-P6, YOLOR-W6 has the same structure but has more parameters than YOLOR-P6. We use the same training setting mentioned above in training the YOLOR-W6 network.

\noindent\textbf{Multi-Model Ensemble}. For each network structure, we first use the scale enhancement strategy mentioned at \ref{CRE} to ensemble the predictions of different scales. Then, according to Table \ref{P6vsW6}, YOLOR-W6 performs well in the vehicle and pedestrian class while YOLOR-P6 performs well in the cyclist class. We choose the vehicle and pedestrian boxes from YOLOR-W6 and gather the cyclist boxes from YOLOR-P6. In this strategy, we can collect advantages from different models.

\begin{table}[h]
\begin{center}
\caption{Experiment on mini-val dataset.}
\begin{tabular}{l|c|c|c|c}
\hline
Method  & Veh. & Ped. & Cyc. & Mean \\
\hline
YOLOR-P6 & 63.5 & 74.9 & \textbf{60.4} &  66.3 \\
YOLOR-W6 & \textbf{64.3} & \textbf{75.3} & 58.9 & 66.2 \\
\hline
\end{tabular}
\label{P6vsW6}
\end{center}
\end{table}

\subsection{Results}
As shown in Table \ref{ablation}, based on the YOLOR detector, we obtained the baseline model and carried out five improvement experiments to steadily improve the results. In the baseline model, we find the small objects are missed and some error results are detected due to difficult annotations. Next, we propose auto data cleaning, which improves the quality of the dataset a lot and increase the performance by 1.30\%; Then, we add multi-scale training and get an improvement by 1.26\%;
So far, our model worked pretty well for large objects. For small objects, we use scale enhancement strategy and get 1.88\% increasing; In addition, we introduce independent threshold-NMS to get better performance. In the last, we ensemble different model which performs better on different class to improve the final performance. At this point, our algorithm’s total performance in mAP has increased from 61.54\% to 66.67\%, an increase of 5.13\%. Details are shown in Table \ref{ablation}.

\begin{table}[h]
\begin{center}
\caption{Ablation study on the mini-val dataset.}
\begin{tabular}{l|c}
\hline
Method  & AP/L2 (\%) \\
\hline
YOLOR Baseline & 61.54 \\
+ Auto Data Cleaning & 62.84\\
+ Multi-Scale Training & 64.10 \\
+ Scale Enhancement & 65.98 \\
+ Independent threshold-NMS & 66.30 \\
+ Model Ensemble & 66.67  \\
\hline
\end{tabular}
\label{ablation}
\end{center}
\end{table}

\section{Conclusion}

In this report, we present a state-of-the-art real-time 2D object detection framework for self-driving scenarios. Specifically, in order to achieve robust detection results, we aggregate the popular one-stage object detectors with the scale enhancement strategy. In addition, we train the models of various input strategies independently, to yield better performance for accurate multi-scale detection of each category. Our overall detection framework achieves the 2nd place in the realtime 2D detection track of the Waymo Open Dataset Challenges.
 
{\small
\bibliographystyle{ieee_fullname}
\bibliography{egbib}
}

\end{document}